\documentclass[letterpaper, 10 pt, conference]{IEEEtran}  

\IEEEoverridecommandlockouts                              

\usepackage{amsmath,amssymb,amsfonts}
\usepackage{algorithmic}
\usepackage{mathtools}
\usepackage{graphicx}
\usepackage{textcomp}
\usepackage{xcolor}
\usepackage{subcaption}
\usepackage{cuted}    
\usepackage{capt-of}  
\usepackage{cleveref}
\usepackage{wrapfig}
\usepackage{adjustbox}   
\usepackage[font=small]{caption}     
\usepackage{multirow}
\usepackage{booktabs,colortbl,xcolor}
\usepackage{subcaption}

\newsavebox{\tablebox}

\begin{document}
\title{\LARGE \bf Constraining Streaming Flow Models for \\ Adapting Learned Robot Trajectory Distributions
}
\author{Jieting Long$^{1}$ \and Dechuan Liu$^{2,3}$ \and Weidong Cai$^{1}$ \and Ian Manchester$^{2,3}$ \and Weiming Zhi$^{1,3,4*}$
\thanks{$^{*}$email: {\tt\small Weiming.Zhi@sydney.edu.au}.}%
\thanks{$^{1}$School of Computer Science, The University of Sydney, Australia.}
\thanks{$^{2}$School of Aerospace, Mechanical and Mechatronic Engineering, The University of Sydney, Australia.}
\thanks{$^{3}$Australian Center For Robotics, The University of Sydney, Australia.}
\thanks{$^{4}$College of Connected Computing, Vanderbilt University, TN, USA.}
}
\maketitle

\begin{strip}
\vspace{-6em}
    \centering    \includegraphics[width=\textwidth]{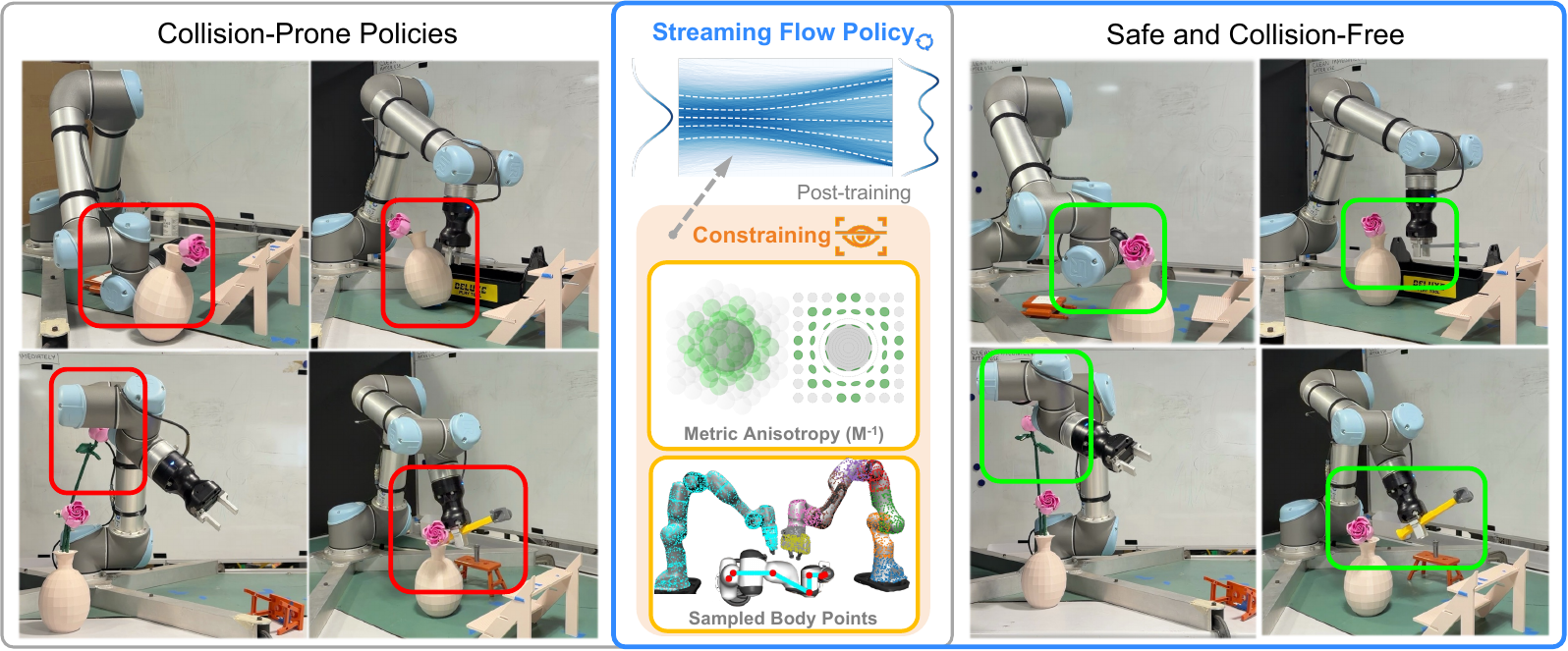}
    \captionof{figure}{Overview of CASF and its effect. \textbf{Left:} collisions produced by the baseline SFP. \textbf{Right:} collision-avoiding behaviours under CASF. \textbf{Middle:} the proposed CASF framework, which introduces geometry- and distance-aware constraints as a post-training module.}
    \vspace{-1em}
    \label{fig:fig1}
\end{strip}

\begin{abstract}
Robot motion distributions often exhibit multi-modality and require flexible generative models for accurate representation. Streaming Flow Policies (SFPs) have recently emerged as a powerful paradigm for generating robot trajectories by integrating learned velocity fields directly in action space, enabling smooth and reactive control. However, existing formulations lack mechanisms for adapting trajectories post-training to enforce safety and task-specific constraints. We propose \emph{Constraint-Aware Streaming Flow} (CASF), a framework that augments streaming flow policies with constraint-dependent metrics that reshape the learned velocity field during execution. CASF models each constraint, defined in either the robot’s workspace or configuration space, as a differentiable distance function that is converted into a local metric and pulled back into the robot’s control space. Far from restricted regions, the resulting metric reduces to the identity; near constraint boundaries, it smoothly attenuates or redirects motion, effectively deforming the underlying flow to maintain safety. This allows trajectories to be adapted in real time, ensuring that robot actions respect joint limits, avoid collisions, and remain within feasible workspaces, while preserving the multi-modal and reactive properties of streaming flow policies. We demonstrate CASF in simulated and real-world manipulation tasks, showing that it produces constraint-satisfying trajectories that remain smooth, feasible, and dynamically consistent, outperforming standard post-hoc projection baselines.
\end{abstract}

\section{Introduction}

Robots operating in complex environments must generate trajectories that are both expressive and reactive, often under multi-modal task distributions. Capturing this variability requires generative motion models capable of representing rich distributions over future trajectories while remaining suitable for real-time control. Recent progress in flexible generative models has enabled robot policies to incorporate complex trajectory distributions, including diffusion-based policies~\cite{janner2022diffuser,chi2023diffusionpolicy,streamingDP} and flow-matching models~\cite{lipman2023flow,fmPolicy2,jiang2025streaming}.

A key drawback of these models lies in their inference procedure, which requires numerically solving an integral over full trajectories, where each step jointly refines all future actions. This limitation was recently addressed by \emph{Streaming Flow Policies} (SFPs)~\cite{lipman2023flow,jiang2025streaming}, which reinterpret the generative process as a continuous-time flow directly in action space rather than in trajectory space. Instead of evolving a trajectory of trajectories, SFP learns a neural velocity field whose integration progressively refines the predicted action sequence. Each flow integration step updates the entire trajectory distribution, allowing actions to be streamed incrementally as they are generated. This formulation provides reduced inference latency compared to diffusion-based sampling, making flow-based generative control practical for closed-loop execution.

However, real-world robots must respect diverse safety and feasibility constraints that may change or be partially unknown at training time. These include joint limits, self-collisions, obstacle avoidance, and workspace boundaries, defined across both configuration and task spaces. Directly integrating the learned flow field can thus lead to infeasible or unsafe motions. To address this, we propose the \emph{Constraint-Aware Streaming Flow} (CASF) framework, which enables the learned ODE to be reshaped online to account for dynamic constraints. CASF models each constraint as a differentiable distance function defining proximity to restricted regions~\cite{Park_2019_CVPR, neuralJsdf_RAL}. These distance functions are used to construct local constraint metrics that deform the geometry of the robot’s action space, scaling and reorienting the underlying flow field according to the local constraint influence. Constraints defined in the robot’s workspace are expressed through their distance fields and \emph{pulled back} via the robot’s kinematics into the control or configuration space, allowing all constraints to be combined coherently. ~\Cref{fig:fig1} illustrates the induced metric ($M^{-1}$) and sampled body points used for metric pullback, together with representative failure and success cases under collisions. The resulting modular metric composition yields a continuously deformed vector field that adapts at runtime to environmental and kinematic changes. This approach enforces constraints directly through the ODE integration process without retraining.

Concretely, this paper makes the following contributions:
\begin{itemize}
    \item Constraint-Aware Streaming Flow (CASF), a framework for reshaping learned streaming flow policies through constraint-dependent metric fields, enabling continuous and differentiable enforcement of safety and feasibility constraints at runtime.
    
    \item A modular constraint composition scheme, where constraints defined in both workspace and configuration space are converted into local metrics, and pulled back through the robot’s kinematics to act coherently within a unified control-space geometry.

    \item Rigorous empirical evaluations demonstrating the capability of CASF to safely enforce constraints while remaining close to behaviours learned via flow-matching.

\end{itemize}

\section{Related Work}
\vspace{0.3em} \noindent\textbf{Motion Policy Generation:} Motion policy generation underpins robotic autonomy, mapping perception directly to closed-loop control. While classical methods primarily relied on model-based planning and optimisation ~\cite{Kooij2018ContextBasedPP,CHOMP,MPC_2017_old}, the field has increasingly shifted toward learning-based paradigms with imitation learning (IL) ~\cite{ravichandar2020recent,ibc, GeometricFab, Diff_templates} as a baseline for cloning expert demonstrations. Recently, generative imitation learning has progressed towards diffusion-based policies ~\cite{janner2022diffuser,chi2023diffusionpolicy,pan2024model,xue2025full}, and most recently to flow-based formulations ~\cite{rmpflow,fmPolicy2,braun2024riemannian,jiang2025streaming}, which retain the expressivity of diffusion while significantly reducing inference latency by replacing iterative denoising with straight-through probability transport along learned flows. Yet, these data-driven policies degrade under distribution shift and frequently collide in unseen environments, as geometric constraints are weakly encoded during training. 
 
\vspace{0.3em} \noindent\textbf{Inference-time Constrained Policy Optimisation:} Constrained policy optimisation formalises motion generation by expressing obstacles as strict constraints, solving for safe trajectories that link the initial and target while respecting kinematic limits. Existing strategies typically fall into two categories. (1) \emph{Trajectory Optimisation and Post-hoc Filtering} rectifies unconstrained rollouts using gradient-based planners (e.g., CHOMP~\cite{CHOMP}, TrajOpt~\cite{trajopt}) or control-theoretic safety filters ~\cite{cbf_intV} grounded in set invariance or reachability analysis . However, these decoupled overrides often induce out-of-distribution actions, suffer from local minima, and incur high computational overhead by separating generation from correction. (2) \emph{Generative Guidance} biases the sampling process directly, employing either soft constraints via classifier-based guidance ~\cite{dhariwal2021diffusion,xiao2023safediffuser} which lack strict feasibility guarantees or hard constraints via projection~\cite{DPwConstarints25} or reflected dynamics (e.g., mirror maps)~\cite{lou2023reflected, liu2023mirror}. Notably, \cite{mishra2025eb} constrains generative diffusion models with progressively enforced barriers, and JM2D~\cite{jointModel_diff} aligns diffusion planners with optimisation-based safety via joint sampling. Our approach (CASF) modulates the generative vector field onto flow models via geometry-aware metric shaping, enforcing safety constraints through smooth, reactive steering. By avoiding discontinuous hard projections or expensive retraining, CASF preserves trajectory continuity and imitation fidelity directly during inference.

\section{Preliminaries: Trajectory Distribution Learning with Flows}

\vspace{0.3em} \noindent\textbf{Flow Matching:} Flow matching~\cite{lipman2023flow} is a continuous generative modeling framework that learns a time-dependent velocity field to transport samples from a simple base distribution to a complex data distribution via an ordinary differential equation (ODE). Given a sample $a(t) \in \mathcal{A}$ evolving under flow time $t \in [0,1]$,
the flow dynamics are defined as:
\begin{align}
\frac{d a(t)}{d t} = v_\theta(a(t), t),
\end{align}
where $v_\theta : \mathcal{A} \times [0,1] \to T\mathcal{A}$
is a neural network parameterising a velocity field. Integrating this ODE transforms an initial distribution $p_0(a)$ into a target distribution $p_1(a)$.

The training objective minimises the discrepancy between the model-predicted velocity $v_\theta(a, t)$ and a target velocity $v^*(a, t)$ constructed from data trajectories:
\begin{align}
\mathcal{L}_{\text{FM}} =
\mathbb{E}_{t \sim \mathcal{U}[0,1],\, a \sim p_t(a)}
\big[\,
\|v_\theta(a, t) - v^*(a, t)\|_2^2
\,\big].
\end{align}
At optimality, the induced marginal distribution of $a(t)$ under $v_\theta$ matches the data distribution at all intermediate timesteps, allowing the model to represent rich, multi-modal densities over continuous variables~\cite{ibc}.

\vspace{0.3em} \noindent\textbf{Streaming Flow Policy:}
The \emph{Streaming Flow Policy (SFP)} simplifies diffusion and flow-matching policies by treating \emph{action trajectories as flow trajectories} \cite{jiang2025streaming}. Traditional diffusion or flow policies generate a \emph{trajectory of trajectories}. That is, they sample a full sequence of actions via a diffusion or flow process that must complete before any robot actions can be executed. This is computationally expensive and introduces latency.

In contrast, SFP operates directly in action space $\mathcal{A}$, defining a history-conditioned velocity field $v_\theta(a, t \mid h)$ that evolves actions according to:
\begin{align}
\frac{d a(t)}{d t} = v_\theta(a(t), t \mid h),
\quad
a(0) \sim \mathcal{N}(a_{\text{prev}}, \sigma_0^2 I),
\end{align}
where $h$ denotes the observation history and $a_{\text{prev}}$
is the most recently executed action.
By incrementally integrating this ODE,
SFP generates a sequence of future actions whose \emph{flow time}
aligns with real execution time, enabling \emph{on-the-fly streaming}
of actions to the robot controller.

During training, stabilising conditional flows are analytically constructed around demonstration trajectories $\xi(t)$ as:
\begin{align}
v_\xi(a, t) = \dot{\xi}(t) - k\, (a - \xi(t)),
\end{align}
whose induced marginals $p_\xi(a \mid t)$ form narrow Gaussian tubes centered around the demonstrations.
The model $v_\theta$ is then trained to match these conditional flows using the conditional flow-matching loss.

The resulting learned velocity field produces per-timestep marginals that match the data distribution, preserving multi-modality while enabling low-latency, closed-loop control suitable for real-time robotic policy execution.

\section{Constraint-Aware Streaming Flow}
The Constraint-Aware Streaming Flow (CASF) framework reshapes the learned velocity field of a Streaming Flow Policy (SFP) to enforce safety and feasibility constraints during execution.  
Rather than retraining or projecting trajectories post hoc, CASF modifies the \emph{geometry} of the flow ODE itself. It treats the robot’s action or configuration space as a deformable manifold whose local geometry adapts near constraint boundaries. As the robot approaches obstacles or joint limits, the local metric stretches unsafe directions and compresses tangential ones, naturally bending or slowing the flow. This is analogous to fluid motion around a solid object. Far from constraints, this metric smoothly reduces to the identity, preserving the nominal behaviour. Through this metric-weighted deformation, CASF embeds constraint awareness directly into the ODE integration process, enabling smooth, real-time adaptation of streaming policies while maintaining their reactive, multi-modal character.

In the rest of this section, we first describe how to construct metrics from distance functions to an infeasible region. We then outline how users can either craft or learn distance functions based on the constraints desired. Finally, we introduce how constraints defined in separate \emph{task-spaces} (e.g. in the robot's workspace and joint space) can be fused together by pulling the constraints to the robot's configuration space.

\begin{figure}[t]
    \centering
    \includegraphics[width=\linewidth]{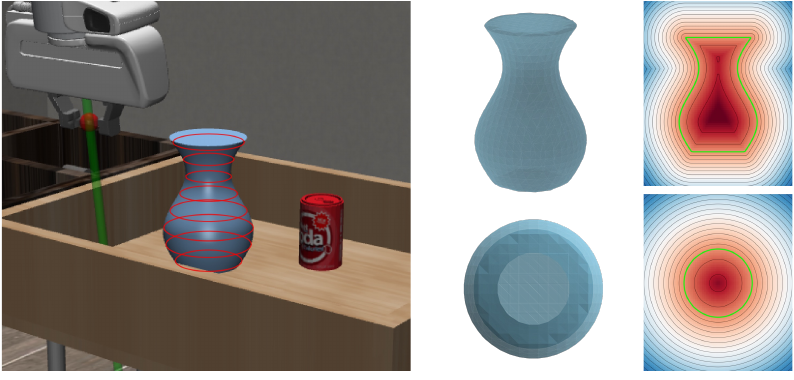}
    \caption{Visualisation of learned neural signed distance fields (Neural SDFs). Left: Robomimic simulation scene with a vase obstacle placed on the table. Middle: zero level-set surface reconstructions of the learned SDF from two viewpoints (camera-aligned and top-down). Right: corresponding 2D SDF slices.}
    \label{fig:neuralSDF}
\end{figure}

\subsection{Distance-Induced Metrics and Metric-Weighted Streaming}
\textbf{Metric Construction:} We define a smooth Riemannian metric that increases motion cost near infeasible or unsafe regions of the action space.
Intuitively, the metric acts as a local ``resistance field'' that slows or diverts motion in the direction of constraint
violation while allowing tangential movement along constraint surfaces. Let $\mathcal{F}$ denote the set of infeasible states, and let $d_k(a) \ge 0$ be differentiable distances from the current
action $a$ to each boundary $\mathcal{F}_k$. With outward unit normals
$n_k(a) = \nabla_a d_k(a) / \|\nabla_a d_k(a)\|$, the local metric is constructed as:
\begin{align}
M(a) = &I
+ \sum_{k} w_k\!\big(d_k(a)\big)
\Big[
\alpha_k\!\big(d_k(a)\big)\, n_k n_k^\top 
\ldots \nonumber
\\ & + \beta_k\!\big(d_k(a)\big)\, (I - n_k n_k^\top)
\Big], \label{eqn:metric_construction}
\end{align}
where $w_k(d)$ is a smooth influence function (e.g., Gaussian), and
$\alpha_k(d)$, $\beta_k(d)$ control the stiffness along and tangent to constraint boundaries. As $d_k(a)\!\to\!\infty$, $w_k(d)\!\to\!0$ and $M(a)\!\to\! I$, so the metric smoothly reverts to the identity far from constraints.

\textbf{Metric-Weight Streaming Flow:} Given the original trained streaming flow $v_\theta(a, t \mid h)$, which represents the learned
velocity field used to generate actions during execution, constraint adaptation is introduced
at inference time by pre-multiplying the velocity with the inverse of the local constraint
metric:
\begin{align}
\frac{d a(t)}{dt}
= \tilde v(a, t \mid h)
= M(a)^{-1} \, v_\theta(a, t \mid h),
\end{align}
where $M(a) \succ 0$ is a smoothly varying, geometry-induced metric constructed from obstacle distance and surface normals, and modulated by a distance-based decay so that shaping vanishes away from constraint boundaries.

Equivalently, $\tilde v$ minimises the metric-weighted projection:
\begin{align}
\tilde v
= \arg\min_{\dot a}
\tfrac{1}{2}\|\dot a - v_\theta\|_{M(a)}^2,
\end{align}
which attenuates motion along constraint normals relative to tangential components, yielding a continuously deformed flow field that enforces constraint-awareness in real time,
without retraining or altering the nominal policy.

\textbf{Connection to Control Barrier Functions (CBFs): }  
The metric-weighted projection formulation can be viewed as a continuous and analytic approximation to safety-filter methods based on Control Barrier Functions (CBFs), which are widely used in constrained robot control. Classical CBF formulations compute a corrected control $\dot a^*$ at each timestep by solving a quadratic program of the form:

\begin{align}
\dot a^* = \arg\min_{\dot a} 
\tfrac{1}{2}\|\dot a - v_\theta(a)\|_2^2
\quad \text{s.t.} \quad g_i(a,\dot a) \ge 0,
\end{align}
where $g_i(a, \dot{a}) \ge 0$ represents a generalised set of feasibility constraints, such as joint limits or actuator saturation, ensuring that the resulting action $\dot{a}^*$ remains within the robot's physical and operational capabilities.

In contrast, CASF embeds an analogous correction directly into the streaming ODE by replacing explicit inequality constraints with a continuously varying metric $M(a)$ whose eigenstructure encodes direction-dependent penalties derived from local geometry. The resulting update $\tilde v = M(a)^{-1} v_\theta(a)$ performs an implicit soft projection of the policy-induced velocity field, approximating the effect of CBF-based safety filtering in closed form without solving an optimization problem at each step. This yields a lightweight, differentiable alternative to CBF-QP filters, maintaining continuous deformation of the velocity field while retaining the interpretability of constrained optimisation.

\subsection{Training Distance Functions for Neural Metric Fields}
To construct constraint-aware metrics, CASF requires smooth distance functions $d_k(a)$ that encode the signed proximity of an action or configuration $a$ to the corresponding constraint boundary $\mathcal{F}_k$. In simple cases, analytic distance functions can be defined directly (e.g., distance to a box, sphere, or joint limit). However, for complex geometries or learned environment representations, these distances can instead be obtained from neural implicit fields trained from samples. ~\Cref{fig:neuralSDF} visualises a learned distance field of a vase in our simulation environment, where no analytical SDF is available, illustrating both the reconstructed surface geometry and corresponding 2D SDF slices.

\textbf{Learning distance functions:} Given a set of sampled points $\{a_i\}_{i=1}^N$ in the robot’s configuration or workspace, with associated ground-truth distances $s_i$ to a constraint region (e.g. obstacles in workspace), we train a neural network $\phi_\psi(a): \mathbb{R}^n \!\to\! \mathbb{R}$ to regress these distances via:
\begin{align}
\mathcal{L}_{\text{dist}}(\psi)=&\mathbb{E}_{(a_i,s_i)}\!\big[\lambda_{\text{mse}}(\phi_\psi(a_i)-|s_i|)^2 \ldots \nonumber\\
&+\lambda_{\text{eik}}(\|\nabla_a\phi_\psi(a_i)\|_2-1)^2\big].
\end{align}
where the first term enforces distance accuracy and the second imposes the Eikonal constraint $\|\nabla_a \phi_\psi(a)\|_2 = 1$ to encourage locally metric-consistent gradients.  
This loss formulation parallels those used in distance function learning for shape reconstruction~\cite{Park_2019_CVPR}, ensuring that $\phi_\psi(a)$ behaves as a smooth distance field in the vicinity of constraint boundaries.

\textbf{Constraint encoding:} Once trained, the neural field $\phi_\psi(a)$ serves directly as the differentiable distance function $d_k(a)$ used for metric construction in~\Cref{eqn:metric_construction}. Its gradient
$\nabla_a \phi_\psi(a)$ defines local surface normals, which are orthogonal to the iso-contour lines visualised in the 2D slices of ~\Cref{fig:neuralSDF}. These normals enable the computation of anisotropic metrics:
\begin{align}
M(a) = &I + w(\phi_\psi(a)) 
\Big[
\alpha(\phi_\psi(a))\, n n^\top \ldots, \nonumber \\ +
&\beta(\phi_\psi(a)) (I - n n^\top)
\Big],
\end{align}
where $n = \nabla_a \phi_\psi(a) / \|\nabla_a \phi_\psi(a)\|$ is the learned constraint normal.  
This yields a smooth neural metric field that remains differentiable.

\subsection{Pullback of Workspace Constraints to C-Space}
\begin{figure}[t]
    \centering
    \includegraphics[width=0.9\linewidth]{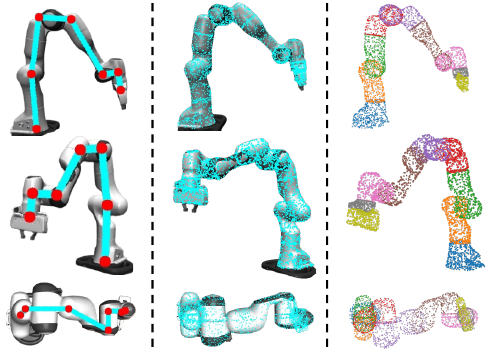}
    \vspace{0.5em}
    \caption{We sample body points on the robot. Left: The joints and links of the manipulator. Middle: dense surface sampling over the full arm mesh. Right: link-wise surface sampling, preserving local geometry per link and used in our pullback-based constraint formulation.}
    \label{fig:bodypoints}
\end{figure}

Robot motion is governed by forward kinematics that map joint configurations to workspace positions.  
Let $q \in \mathcal{C} \subset \mathbb{R}^n$ denote the configuration vector, and $x = f(q) \in \mathcal{W} \subset \mathbb{R}^3$ the corresponding workspace point. The differential relation:
\begin{align}
    \dot{x} = J(q)\, \dot{q},
\end{align}
uses the Jacobian $J(q) = \partial f(q) / \partial q$ to map joint velocities to workspace velocities.  
This extends naturally to multiple body points $f_i(q)$, as illustrated in \Cref{fig:bodypoints}, with multiple Jacobians $J_i(q)$, allowing proximity and collision constraints to be enforced across the robot structure.

\begin{table*}[t]
\centering
\begin{minipage}[b]{0.3075\textwidth}

    \centering
    \begin{adjustbox}{width=\textwidth}
    \begin{tabular}{ll|rrr}
    \toprule
       & Obs & S.R. & G.C. & A.P.L \\

       \midrule\midrule
       \multirow{2}{*}{NoShaping} & State & 0.479 & 0.836 & 193 \\ 
            & Visual & 0.469 & 0.819   & 193\\\arrayrulecolor{gray!40}\cmidrule(lr){2-5}\arrayrulecolor{black} 
       \multirow{2}{*}{Hard-Projection} & State & 0.480 & 0.796 & 194 \\ 
            & Visual & 0.400 & 0.711   & 187\\\arrayrulecolor{gray!40}\cmidrule(lr){2-5}\arrayrulecolor{black} 
       \multirow{2}{*}{CBF} & State & 0.780 & 0.899 & 183 \\ 
            & Visual & 0.600 & 0.805 & 185\\\arrayrulecolor{gray!40}\cmidrule(lr){1-5}\arrayrulecolor{black}  
       \multirow{2}{*}{\textbf{CASF (Ours)}} & State  & \textbf{0.813} & \textbf{0.961} & \textbf{168} \\ 
            & Visual & \textbf{0.816} & \textbf{0.941} & \textbf{157}\\
    \addlinespace \hline 
    
    \bottomrule
    \end{tabular}
    \end{adjustbox}
    \caption{Push-T: At the last step of each episode, we measure the \textbf{coverage} (G.C.) of the block over the goal area, the \textbf{success rate} (S.R.), where success is defined as a higher coverage than the minimum in the demonstration dataset, and the average trajectory \textbf{length} (A.P.L.). }
    \label{table_pusht}
    \end{minipage}\hfill
    \begin{minipage}[b]{0.67\textwidth}
    \begin{adjustbox}{width=\textwidth}
    \begin{tabular}{ll|rrrrrrrrr}
    \toprule
        & Metric & Line & Khamesh & N-Shape & Sine & R-Shape & S-Shape & W-Shape & Worm & Z-Shape \\
       \midrule\midrule
    \multirow{3}{*}{SFP} & Masked F.D.  & NA & NA & NA & NA & NA & NA & NA & NA & NA \\ 
    & M.P.D. & 0.159 & 0.219 & 0.131 & 0.237 & 0.091 & 0.213 & 0.219 & 0.169 & 0.270  \\ 
    & IntViolation & 14.526 & 18.921 & 6.516 & 8.399 & 9.245 & 8.738 & 6.714 & 7.186 & 9.768   \\

    \addlinespace \hline \addlinespace
                        
    \multirow{3}{*}{Projection} & Masked F.D. & 0.139 & 0.163 & 0.156 & 0.294 & 0.077 & 0.199 & 0.385 & 0.285 & 0.418  \\ 
         & M.P.D. & 0.001 & 0.002 & 0.002 & 0.003 & 0.001 & 0.003 & 0.003 & 0.002 & 0.003  \\
 & IntViolation & 0.054 & 0.088 & 0.039 & 0.157 & 0.028 & 0.118 & 0.094 & 0.055 & 0.168  \\ 
    
    \addlinespace\hline\addlinespace
    
    \multirow{3}{*}{CBF} & Masked F.D & \textbf{0.142} & 0.142 & 0.119 & \textbf{0.205} & 0.075 & 0.267 & 0.383 & 0.274 & 0.370 \\ 
         & M.P.D. & 0.000 & 0.001 & \textbf{0.000} & \textbf{0.000} & \textbf{0.000} & \textbf{0.000} & \textbf{0.000} & \textbf{0.000} & \textbf{0.000}  \\
 & IntViolation & \textbf{0.000} & 0.037 & \textbf{0.000} & \textbf{0.000} & \textbf{0.000} & \textbf{0.000} & \textbf{0.000} & \textbf{0.000} & \textbf{0.000} \\ 
    
    \addlinespace\hline\addlinespace
    
    \multirow{3}{*}{\textbf{CASF (Ours)}} & Masked F.D & \textbf{0.051} & 0.157 & \textbf{0.117} & 0.211 & \textbf{0.054} & \textbf{0.093} & \textbf{0.186} & \textbf{0.080} & \textbf{0.164}  \\ 
         & M.P.D. & \textbf{0.000} & \textbf{0.000} & \textbf{0.000} & \textbf{0.000} & \textbf{0.000} & \textbf{0.000} & \textbf{0.000} & \textbf{0.000} & \textbf{0.000}  \\
 & IntViolation & \textbf{0.000} & \textbf{0.000} & \textbf{0.000} & \textbf{0.000} & \textbf{0.000} & \textbf{0.000} & \textbf{0.000} & \textbf{0.000} & \textbf{0.000} \\ 
    
    \addlinespace \hline 
    
    \bottomrule
    \end{tabular}
    \end{adjustbox}
    \caption{Quantitative performance comparison of constraint-handling methods (raw SFP - no shaping, Hard-Projection, CBF, and CASF) on LASA obstacle-avoidance tasks \cite{lasa2011}. Lower is better for all metrics: Masked Fréchet Distance (Masked F.D.)~\cite{Eiter1994ComputingDF}, Maximum Penetration Depth (M.P.D.)~\cite{maxPenDist}, and Integral Violation (IntViolation)~\cite{cbf_intV}. }
    \label{tab:table_lasaObstacle}
\end{minipage}
\vspace{-1.5em}
\end{table*}

\textbf{Metric pullback:}  
Workspace constraints are often represented as distance fields $\phi_i(x)$ defining proximity to obstacles or boundaries in $\mathcal{W}$, whereas streaming flow policies operate in configuration space $\mathcal{C}$.  
To align these spaces, each workspace metric $M_{x,i}(x)$ is \emph{pulled back} via:
\begin{align}
    M_{q,i}(q) = J_i(q)^\top\, M_{x,i}\big(f_i(q)\big)\, J_i(q),
\end{align}
ensuring that penalised directions in workspace correspond to their joint-space counterparts under the robot’s kinematics. Intuitively, if a workspace direction is heavily penalised (e.g., moving into an obstacle), the corresponding joint-space directions are also penalised in $M_{q,i}(q)$.

\textbf{Metric fusion:}  
To account for collisions along the full robot geometry, we aggregate the pulled-back metrics of each body point sampled into a composite configuration-space metric $M_{\text{fused}}(q)$. The fusion is anchored with an identity-based baseline to ensure the metric remains well-conditioned in free space:
\begin{align}
    M_{\text{fused}}(q) = I + \sum_i w_i(q)\, M_{q,i}(q),
\end{align}
where $w_i(q) = \exp\!\big(-\kappa_i d_i(q)^2\big) \in [0,1]$ is an SDF-based activation weight, modulating each constraint’s influence with the distance $d_i(q)$ and decay rate $\kappa_i$. 

Near active constraints, their metrics dominate; far away, as the robot moves into free space, $w_i(q) \to 0$ and $M_{\text{fused}}(q) \to I$.  During execution, the nominal streaming flow $v_{\theta}$ is deformed into a constraint-aware velocity field by solving the metric-weighted projection: 
\begin{align}
    \frac{d q(t)}{dt}
    = M_{\text{fused}}(q)^{-1}\, v_\theta(q, t \mid h),
\end{align}
blending the effects of joint-space and workspace constraints into a unified, smooth constraint-aware flow.

\begin{figure}[t]
    \centering

    \begin{subfigure}{\linewidth}
        \centering
        \includegraphics[width=\linewidth]{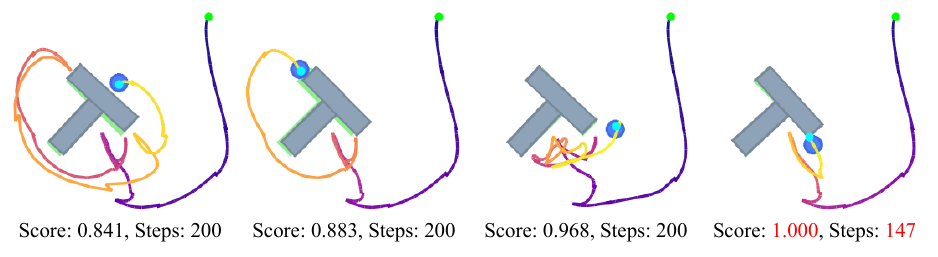}
        \caption{pushTstate}
    \end{subfigure}


    \begin{subfigure}{\linewidth}
        \centering
        \includegraphics[width=\linewidth]{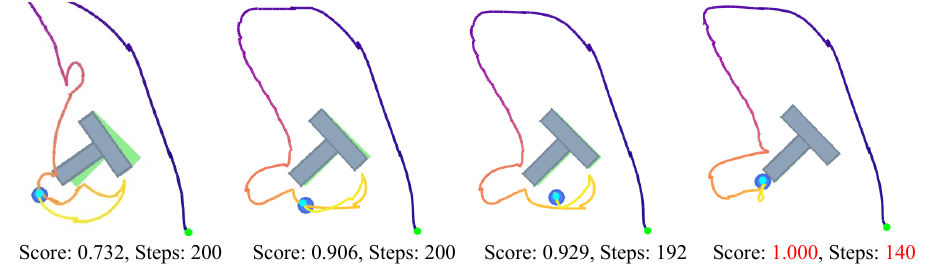}
        \caption{pushTvisual}
    \end{subfigure}

    \caption{Comparison of representative rollouts under different constraint-handling strategies: No Shaping(left), Hard-Projection (left-second), CBF (right-second), and CASF (right). Each panel reports the final task score and number of steps.}
    \label{fig:figure_pusht}
\end{figure}

\begin{figure*}[t] 
\centering 
\includegraphics[width=\linewidth]{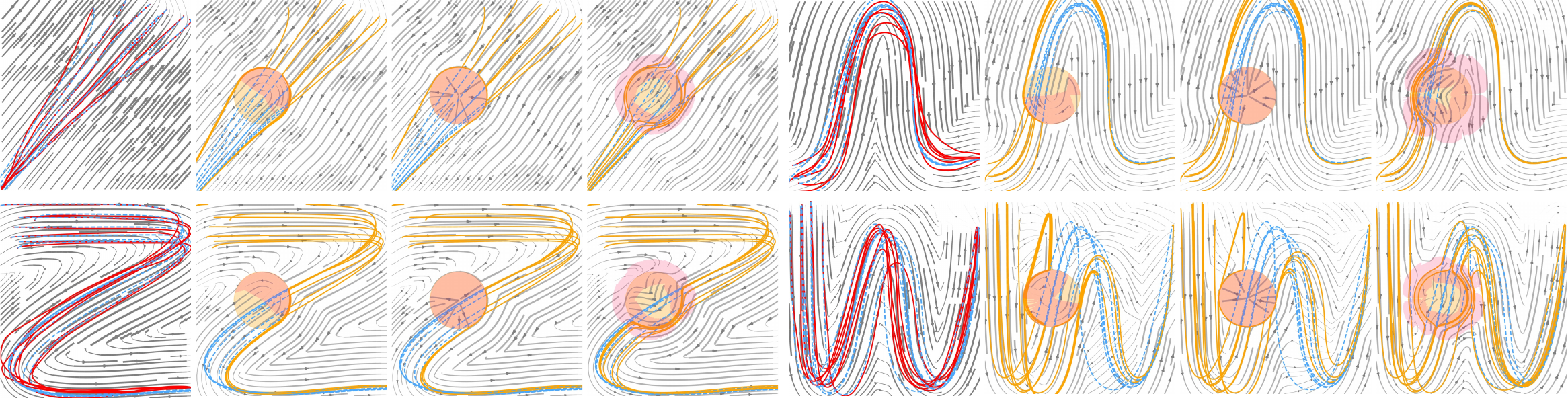} \caption{Qualitative evaluation of reactive obstacle avoidance on four LASA tasks (Line, Worm, Z-shape, and W-shape). Each task compares the raw policy, Hard-Projection, CBF, and CASF. Grey streamlines visualize the induced velocity fields, while orange curves show rollout trajectories. The blue dashed curve denotes the unshaped predicted trajectory, and the red curve in the first column indicates the ground-truth demonstration. Transparent reddish overlays highlight the correction masks introduced by each shaping method.} \label{fig:lasaObstacle}
\vspace{-1em}
\end{figure*}

\begin{figure*}[t] \centering \includegraphics[width=\linewidth]{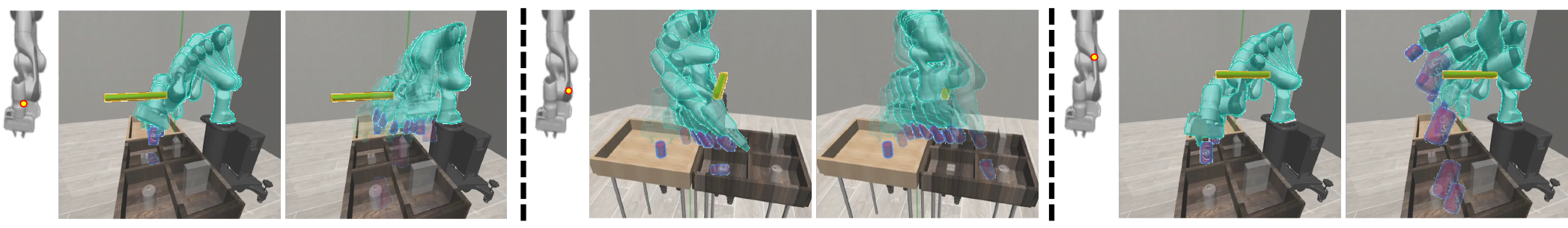} \caption{Whole robot collision avoidance. Each group demonstrates avoidance behaviour induced by collisions at body points along the links (yellow dot markers). Left: rollouts of the baseline imitation policy. Right: CASF with collision-avoidance over the body of the robot.} \label{fig:robomimic_wholebody} \vspace{-1.75em} \end{figure*} 

\begin{figure*}[t] \centering \includegraphics[width=\linewidth]{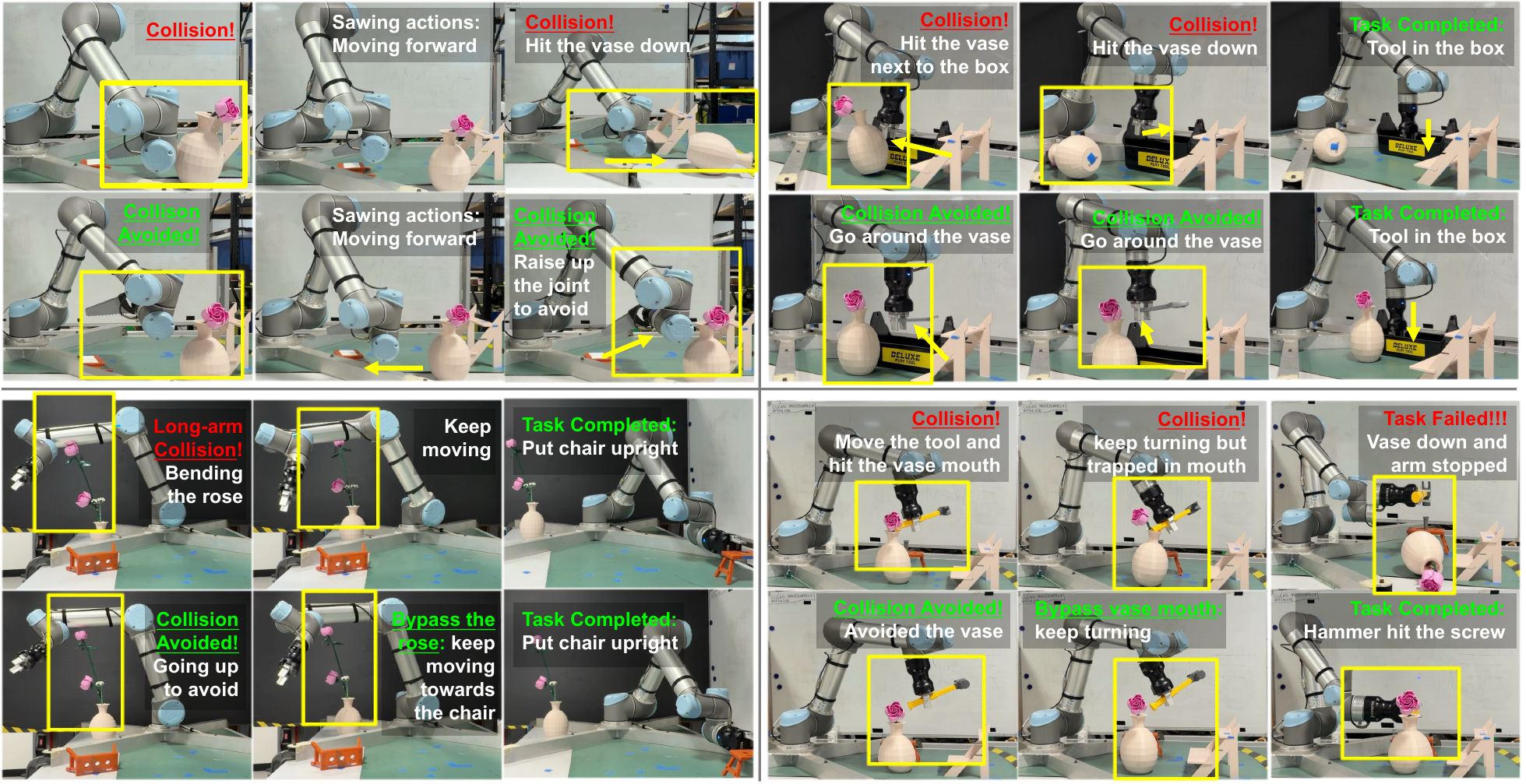} \caption{CASF deployed in the real-world. Across four manipulation tasks, policies without CASF collide with obstacles (top), whereas CASF enables collision-aware motion generation that avoids contact while completing the tasks (bottom).} 
\label{fig:realRobot_result} \vspace{-1.5em} \end{figure*} 

\begin{figure}[t] \centering \includegraphics[width=\linewidth]{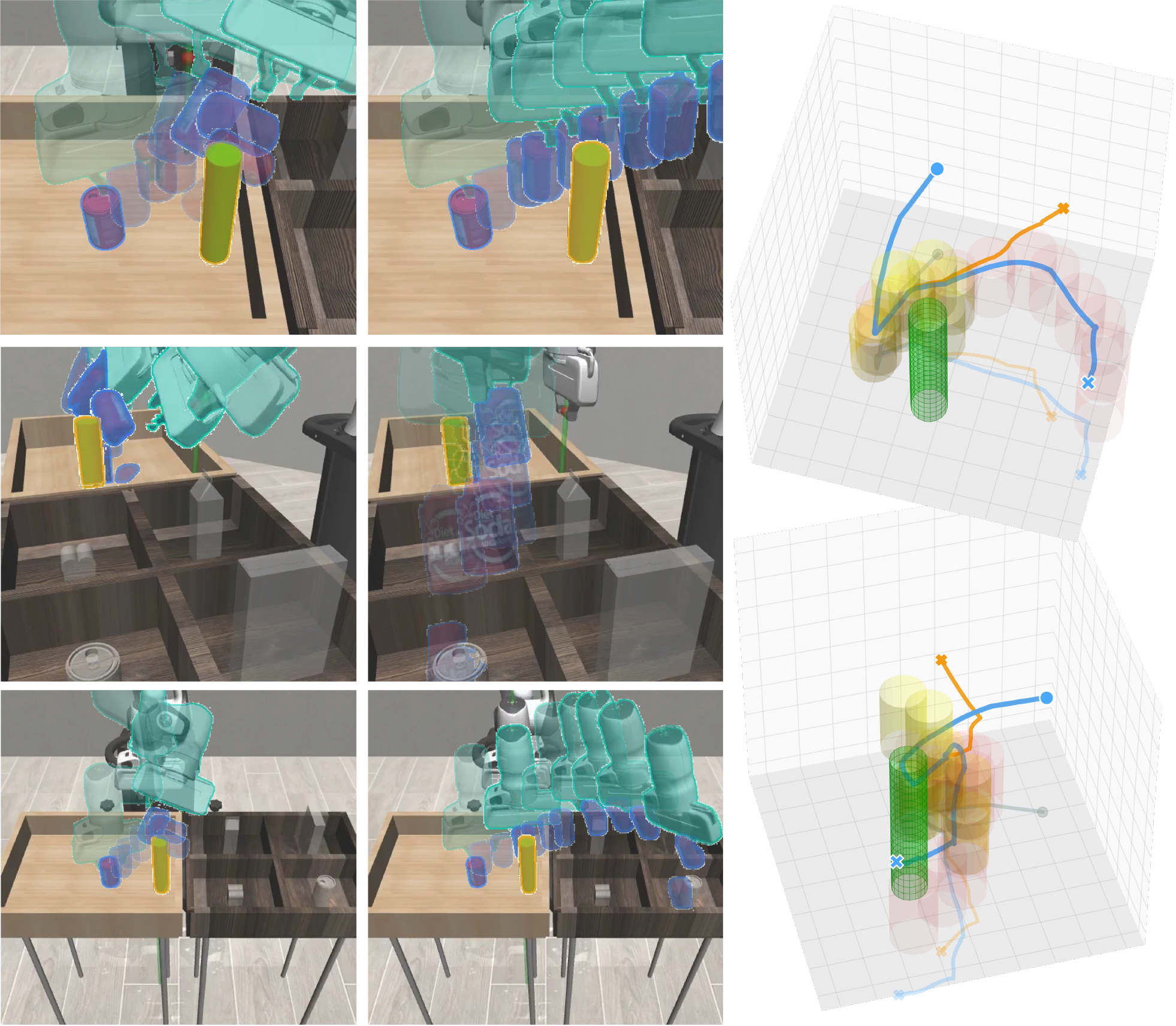} \caption{
Workspace collision avoidance in the 3D Robomimic environment. Left: policy executions with obstacles (green cylinders), with translucent robot and object meshes illustrating the swept motion volume (first column: baseline policy; second column: CASF). Right: 3D end-effector trajectories, comparing raw policy rollouts (orange) with CASF-shaped trajectories (blue)} \label{fig:robomimic_qual} \end{figure} 

\section{Experiments}

We rigorously investigate the capabilities of Constraint-Aware Streaming Flow Policy (CASF), examining its ability to enforce robust collision avoidance. CASF is benchmarked against hard projection and Control Barrier Functions (CBFs)~\cite{cbf_intV} filtering. The empirical evaluations are conducted both in simulation and in the real-world on a Universal Robot (UR) manipulator. When training underlying streaming flow policies, we follow the hyperparameters and setup in \cite{jiang2025streaming}.

\vspace{0.3em} \noindent\textbf{Constraining the workspace:} To first evaluate CASF’s ability to enforce geometric constraints directly, we assess its global and local constraint awareness in the PushT~\cite{implicitBC} and LASA~\cite{lasa2011} environments, respectively, and further validate it in a 3D RoboMimic~\cite{robomimic2021} setting. All experiments adopt a closed-loop, receding-horizon evaluation protocol, in which predicted velocity chunks are integrated forward and resulting states are fed back for subsequent actions.

To assess policy fidelity under large-scale workspace constraints, we evaluate performance in the 2-D PushT environment. Policies are trained via imitation learning using 200 demonstrations for both state- and image-conditioned variants and evaluated over 50 closed-loop episodes. Quantitative evaluations are conducted via the following metrics:
\begin{itemize}
\item \textit{Success Rate (S.R.)}: the proportion of episodes in which the object reaches the goal region, summarising overall task completion under a success criterion;
\item \textit{Goal Coverage (G.C.)}: the averaged fraction of the goal region covered by the object at the final timestep, reflecting how well the task objective is ultimately achieved;
\item \textit{Average Path Length (A.R.L.)}: the mean length of the executed end-effector trajectory, indicating execution efficiency and detour severity under constraint enforcement.
\end{itemize}

Quantitative results (\Cref{table_pusht}) show that CASF consistently outperforms the unconstrained SFP baseline and projection-based filters, achieving higher Success Rate and Goal Coverage while improving operational efficiency with a ~15\% reduction in path length. Qualitative inspection (\Cref{fig:figure_pusht}) confirms that, unlike hard-projection and CBF-style corrections, which often induce oscillatory drift or blocking effects due to discontinuous or normal-only adjustments, CASF’s continuous metric shaping imparts global geometric awareness; this allows the policy to anticipate and deflect from boundaries smoothly, maintaining safer and more direct routes while preserving nominal behaviour in free space.

To further investigate the behaviour under localised reactive constraints, we conduct experiments on the LASA benchmark \cite{lasa2011}. We report three quantitative metrics:
\begin{itemize}
    \item \textit{Masked Fréchet Distance (Masked F.D.)}~\cite{Eiter1994ComputingDF} measures trajectory deviation, isolating how much the global shape changes from trajectories of the unshaped policy;
    \item \textit{Maximum Penetration Depth (M.P.D.)}~\cite{maxPenDist} reports the worst-case constraint violation, capturing peak failures;
    \item \textit{Integral Violation (IntViolation)}~\cite{cbf_intV} accumulates violations over time, reflecting the overall severity and duration of unsafe behaviour.
\end{itemize}

As shown in \Cref{tab:table_lasaObstacle}, CASF achieves near-zero safety violations across all tasks, while consistently attaining the lowest Masked F.D., demonstrating that it preserves global trajectory structure substantially better than projection-based baselines while maintaining strict constraint satisfaction. While the unshaped SFP exhibits substantial violations (large M.P.D. and IntViolation), indicating frequent and deep obstacle intrusions. Hard-Projection and CBF both enforce strict safety, driving M.P.D. and IntViolation close to or equal to zero, but at the cost of markedly increased Masked F.D., revealing significant distortion of the nominal trajectory due to abrupt or piecewise corrections. Note that Masked F.D. for the SFP baseline is reported as NA, since the metric measures distortion induced by shaping relative to the raw policy and is thus not meaningful when no shaping is applied. Qualitative results (\Cref{fig:lasaObstacle}) corroborate this finding: CASF induces fluid, anticipatory deviations that minimally perturb the global trajectory shape, in contrast to the abrupt directional changes or piecewise-linear detours produced by hard-projection and CBF corrections. Correction mask overlaid highlights this distinction: CASF induces a spatially graded, distance-modulated shaping of the velocity field, whereas projection-based methods apply discontinuous or boundary-concentrated adjustments.

Further qualitative validation in the 3D Robomimic environment (\Cref{fig:robomimic_qual}) corroborates these findings within a higher-dimensional workspace. The overlaid trajectories and translucent body volumes demonstrate how CASF produces smooth, anticipatory deviations of the end-effector motion around cylindrical obstacles while preserving a coherent approach toward the task goal. In contrast, baseline methods (top row) result in task failure, succumbing to collisions or deadlocks near the obstacle. This performance highlights how the induced Riemannian metric effectively warps the 3D vector field, guiding the end-effector along a fluid, collision-free trajectory without sacrificing task progress.

Overall, these results demonstrate that CASF enforces workspace-level safety without sacrificing trajectory fidelity or efficiency. By reshaping the local geometry of the workspace through a distance-modulated Riemannian metric, CASF enables anticipatory avoidance behaviour that preserves the expressive structure learned by the streaming flow policy.

\vspace{0.3em} \noindent\textbf{Whole Robot Collision Avoidance:} Under the 3D Robomimic manipulation environment, we next evaluate CASF in a more challenging setting that requires whole-robot collision avoidance in high-dimensional configuration space, where obstacles are introduced into the workspace, positioned to cover intermediate links of the manipulator (e.g., elbow, forearm) rather than the end-effector. These scenarios induce collisions that cannot be resolved by task-space corrections alone. Here, we sample up to 15 body points on each of the robot links to enable whole-robot collision avoidance. \Cref{fig:robomimic_wholebody} presents qualitative rollouts under multiple obstacle placements, each targeting a distinct region of the arm (First Column). Baseline policies, which only consider end-effector safety, consistently fail (Second Column): when the end-effector itself is not threatened, task-space safety filters remain inactive, causing intermediate links to drive directly into obstacles. In contrast, CASF (Third Column) successfully adapts the motion by redistributing avoidance behaviour across relevant joints. This produces smooth, coordinated rotations that steer the threatened link clear of the hazard while maintaining task progression.

The overlaid translucent body volumes in \Cref{fig:robomimic_wholebody} illustrate the efficacy of configuration-space metric pullback. By defining distance-modulated metrics at multiple control points along the kinematic link and aggregating their contributions actively via the Jacobian, CASF induces joint-level velocity shaping that reflects the global geometry of the arm–obstacle interaction. Consequently, avoidance emerges as a cohesive whole-robot response: upstream joints adjust preemptively to retract the threatened link, exploiting the kinematic null space to ensure safe clearance without disrupting the end-effector's path toward the goal. Importantly, these corrections remain continuous and task-consistent, avoiding the abrupt halting or oscillatory behaviour commonly observed with projection-based filtering methods.

Collectively, these results demonstrate that CASF extends beyond end-effector safety to enable robust, anticipatory whole-robot collision avoidance in complex 3D manipulation scenarios. By shaping the policy’s velocity field directly in configuration space through metric pullback, CASF preserves the expressive structure of the learned motion while ensuring safety under diverse, previously unseen obstacle interactions.

\vspace{0.3em} \noindent\textbf{Real Robot Execution:} We further assess the performance of CASF on a real robotic platform using four representative manipulation tasks, each featuring distinct obstacle placements that induce contacts with different parts of the arm. The evaluated tasks are:

\begin{enumerate}
\item \textbf{Tidy Up:} Picking up a fallen chair and placing it upright on the target area, with an obstacle blocking the end-effector’s direct path to the goal.
\item \textbf{Pick \& Place:} Retrieving a wrench from the tool shelf and placing it into a storage box, with an obstacle positioned to interfere with the upper arm along the reaching motion.
\item \textbf{Sawing:} Grasping a hand saw from the tool shelf and executing sawing motions, with an obstacle repeatedly contacting the wrist during the action.
\item \textbf{Hammer Hitting:} Picking up a hammer from the tool shelf and hammering a nail, where collisions occur right after grasping the tool.
\end{enumerate}

For each task, we collect 50 demonstration trajectories for each task and train neural SDFs of the environments. We visualise qualitative comparisons between executions with and without collision avoidance in~\Cref{fig:realRobot_result}, illustrating the effectiveness of CASF under different obstacle configurations. Across all tasks, CASF consistently enables collision-aware motion generation under diverse contact scenarios, demonstrating robust obstacle avoidance behaviours while preserving task completion. These results highlight the ability of CASF to generalise across heterogeneous manipulation skills and obstacle configurations on a real robotic platform, despite being trained on relatively small datasets.

\section{Conclusions and Future Work}

We introduced Constraint-Aware Streaming Flow (CASF), a constraint-aware shaping mechanism for streaming flow policies that embeds geometric safety constraints directly into the policy’s inference-time dynamics. By coupling signed distance fields with metric shaping of streaming flow dynamics, CASF provides a principled mechanism to steer learned behaviours away from obstacles and boundaries without altering the underlying policy distribution. Experiments in both simulation and real-robot settings show that CASF produces smooth, stable trajectories that remain faithful to demonstrations.

Future work will explore integrating perception-driven distance fields to support deployment in unstructured scenes. A key theoretical extension is to relax the symmetry of the shaping metric by considering direction-dependent or non-reversible geometric structures (e.g., Finsler or Randers metrics), which can introduce asymmetric modulation of the flow field and explicitly bias motion away from constraint boundaries rather than merely attenuating velocities along surface normals. Such formulations may provide stronger guarantees in tight, non-convex corridors and near-contact scenarios.

\bibliographystyle{IEEEtran}
\bibliography{ref}

\end{document}